\documentclass{article}

\usepackage[preprint]{neurips_2026}

\usepackage[utf8]{inputenc}
\usepackage[T1]{fontenc}
\usepackage{hyperref}
\usepackage{url}
\usepackage{bbm}

\usepackage{booktabs}
\usepackage{amsfonts}
\usepackage{amssymb}
\usepackage{amsmath}
\usepackage{nicefrac}
\usepackage{microtype}
\usepackage{xcolor}
\usepackage{graphicx}
\usepackage{multirow}
\usepackage{amsthm}
\newtheorem{definition}{Definition}
\title{One Policy, Many Styles: Where to Aggregate Preferences for Steerable Humanoid Control?}

\author{
  Ziqian Huang
\And
  Yuzhuo Chen
\And
  Shimin Zhang
\AND
  Junyang He
\And
  Xinyue Annie Yang
\And
  Animesh Garg
\AND
  \normalfont College of Computing\\
  Georgia Institute of Technology
}

\begin{document}
\maketitle

%===============================================================================

\begin{abstract}

Modern robotic agents are increasingly expected to support steerable control, where a single policy can adjust its behavioral style at runtime under changing user preferences. Preference-conditioned reinforcement learning provides this interface, but conditioning alone does not ensure that a requested behavior change is strong, does not leak into unrelated behaviors, and does not compromise task performance. Our evaluation combines Preference Reward Responsiveness (PRR) with coverage and utility metrics. We use these measures to compare reward-level, advantage-level, loss-level, and PCGrad-based aggregation across two humanoid-control frameworks. No aggregation level performs best across all environments and criteria. Experiments show that aggregating objectives early produces a wider range of behavioral responses but greater unintended changes in other objectives, while aggregating them later improves control precision at some cost to responsiveness.
% Which trade-off is preferable depends on the environment and its underlying task demands.
% These results show that steerability cannot be captured by a single aggregate score and motivate training mechanisms that adapt how objective signals are combined.
\end{abstract}

% Keywords are omitted because the NeurIPS 2026 style does not define a
% \keywords command.

%===============================================================================

\section{Introduction}
\label{sec:intro}

    Robotic and embodied agents are increasingly expected to operate under changing user preferences and behavioral constraints. A household robot may need to prioritize speed in one setting, energy efficiency in another, or smooth and safe motion around humans. Rather than training separate policies for every trade-off, an ideal system should support steerable control, where behavior can be adjusted at runtime through a compact and interpretable preference interface.

Multi-objective reinforcement learning (MORL) provides a natural framework for this problem by representing behavior through multiple reward dimensions and conditioning a policy on a preference vector which specifies desired objective trade-offs \citep{basaklar2023pdmorl,cmorl2025}. This enables one policy to expose a continuous control interface at test time, but conditioning alone does not guarantee steerability. Increasing one preference may produce little change in the intended behavior, produce unintended behaviors, or compromise the essential tasks that the robot must continue to perform.

This distinction is not captured by aggregate return or Pareto-coverage measures alone~\citep{molins2026controllability}.
 A useful evaluation should answer three questions: Does the policy respond strongly when a preference changes? Does the change unintentionally affect other behaviors? Does the policy still perform the task reliably without more failures? We therefore formulate preference-conditioned steerability as a multidimensional evaluation problem and introduce a unified suite based on the Preference Reward Responsiveness (PRR) matrix, Self-PRR, Cross-PRR, Selectivity, and Rank Accuracy. These metrics complement standard utility measures by exposing both intended responsiveness and cross-objective leakage \citep{molins2026controllability}.

pWe compare reward, advantage, and loss-level scalarization within PPO~\citep{schulman2017ppo} across two independently implemented humanoid-control systems, ProtoMotions~\citep{tessler2025protomotions} and MimicKit~\citep{peng2025mimickit}, to test whether the findings generalize across implementations. Within ProtoMotions, we additionally test PCGrad-based gradient surgery~\citep{yu2020gradient} as a later aggregation point that resolves objective conflicts directly in gradient space. Reward-level aggregation achieves the widest controllable range in ProtoMotions, whereas later aggregation achieves higher hypervolume in fixed-direction MimicKit. Its weaker cross-objective precision persists across both frameworks, but these experiments do not isolate the cause of their different coverage rankings. A preliminary check in MetaDrive \citep{li2023metadrive}, a driving domain, shows a broadly similar pattern.

Our contributions are:
\begin{enumerate}
    \setlength{\itemsep}{0pt}
    \item We formulate preference-conditioned robot steerability as a multidimensional evaluation problem that separates responsiveness, monotonicity, selectivity, task preservation, and failure behavior.
    \item We present a framework that trains preference-conditioned policies under different PPO-based aggregation strategies and evaluates their steerability using the suite above, validated across two independently implemented humanoid-control systems.
    \item We show that the payoff from aggregating preferences early depends on the experimental setting: coverage and utility rankings differ across frameworks, while later aggregation consistently yields higher selectivity and lower cross-objective leakage.
\end{enumerate}

Code is available at: \url{https://anonymous.4open.science/r/Pref-Align-RL-214C/}.

%===============================================================================
\section{Background and Related work}

Preference-conditioned MORL trains a single policy over preference vectors so that it can represent multiple objective trade-offs at runtime \citep{pdmorl2023,d3po2026}. Recent work applies this to humanoid control through a mixture-of-experts architecture \citep{li2026pchc}, but fixes a single aggregation point and does not measure cross-objective leakage. We instead study how the aggregation point affects whether changing a policy's preference input predictably changes its behavior.

In the LLM literature, Constitutional AI shows how written principles can guide behavior through self-critique and AI feedback \citep{bai2022constitutional}. The constitution shapes training rather than serving as a continuous runtime control. SteerLM enables runtime steering by conditioning generation on desired attribute values, while multi-objective alignment methods combine rewards or model predictions to trade off properties such as helpfulness, safety, and coding ability \citep{dong2023steerlm,wang2024combining,shi2024mod}. Recent methods extend this idea to fine-grained value priorities and efficient adaptation across a Pareto frontier of user preferences \citep{guo2025couple,li2026hoe}. Together, these works motivate an explicit interface for controlling behavior.

The level at which objective signals are combined during training can affect how well each preference controls behavior. The level at which objective signals are combined during training can affect how well each preference controls behavior. Gradient-surgery methods such as PCGrad address objective conflicts by modifying per-objective gradients before they are combined \citep{yu2020gradient}. Preference-conditioned MORL is usually evaluated with Pareto coverage and utility, which do not directly separate intended response from cross-objective leakage. Recent controllability work begins to expose this limitation \citep{molins2026controllability}. We extend this perspective to embodied control by comparing reward, advantage, and loss-level combination and evaluating responsiveness, monotonicity, selectivity, task preservation, and failure behavior across two robot-control settings.

%===============================================================================

% \section{Problem Statement}

%===============================================================================

\section{Preliminary}

\subsection{Preference-Conditioned Multi-Objective Reinforcement Learning}

We model the MORL setting as a multi-objective Markov decision process with vector reward
\[
\mathbf{r}_t =
\big[r_t^{(1)}, \ldots, r_t^{(K)}\big] \in \mathbb{R}^K,
\]
where each dimension corresponds to one objective \citep{hayes2022practical,felten2023toolkit}. A preference vector \(\mathbf{w} \in \Delta^{K-1}\) specifies the desired trade-off among these objectives. A common scalar utility is the linear scalarization
\[
r_t^{\mathrm{scalar}} = \mathbf{w}^{\top}\mathbf{r}_t.
\]

In preference-conditioned MORL, the goal is to train one policy
\[
\pi_\theta(a \mid s, \mathbf{w})
\]
that can be steered by changing \(\mathbf{w}\) at test time. The same policy parameters are shared across preferences, while the input \(\mathbf{w}\) specifies the desired behavior.

Because objectives can conflict, MORL evaluates policies by the trade-offs they recover rather than by a single reward value. A solution is Pareto-dominated if another solution is at least as good in every objective and better in at least one. The Pareto front is the set of non-dominated solutions. In our experiments, we report hypervolume (HV) and expected utility (EU), which measure Pareto-front quality and average utility under sampled preferences, respectively~\citep{hayes2022practical,felten2023toolkit}; formal definitions and our estimation procedure are given in Appendix~\ref{app:pareto_metrics}.

%===============================================================================

\section{Method}
\subsection{Aggregation Strategies}
\label{sec:method_aggregation}

All four methods share the same actor-critic architecture, observation space, and per-objective reward vector $\mathbf{r}_t = [r_t^{(0)}, r_t^{(1)}, \ldots, r_t^{(K)}]$, where $r_t^{(0)}$ is the fixed-weight task reward and $r_t^{(1)}, \ldots, r_t^{(K)}$ are the $K$ preference-controlled style rewards. We write the combined per-step coefficient for objective $k$ as
\[
c_k(\mathbf{w}) =
\begin{cases}
w_{\mathrm{task}} & k = 0 \\
w_{\mathrm{pref}}\, w_k & k \geq 1,
\end{cases}
\]
where $w_{\mathrm{task}}$ is fixed and $\mathbf{w} = (w_1,\ldots,w_K) \in \Delta^{K-1}$ is the sampled preference. Let $\rho_t = \pi_\theta(a_t\mid s_t,\mathbf{w}) / \pi_{\theta_{\mathrm{old}}}(a_t\mid s_t,\mathbf{w})$ denote the importance ratio and $L^{\mathrm{clip}}(A) = \min\big(\rho_t A,\ \mathrm{clip}(\rho_t, 1{-}\epsilon, 1{+}\epsilon)\, A\big)$ the standard PPO~\citep{schulman2017ppo} clipped surrogate for a scalar advantage $A$. All critics are also conditioned on $\mathbf{w}$.

The four methods differ only in \emph{how late} the $K{+}1$ objectives are collapsed into the single scalar signal that ultimately updates the actor, forming a continuum from earliest to latest aggregation.

\paragraph{Reward-level.}
Objectives are combined at the reward itself, before any advantage estimation:
\[
r_t = \sum_{k=0}^{K} c_k(\mathbf{w})\, r_t^{(k)}.
\]
A single scalar critic $V_\phi(s,\mathbf{w})$ is trained on this scalarized reward, standard GAE~\citep{schulman2016gae} produces one advantage $A_t$, and the actor is updated with $L^{\mathrm{clip}}(A_t)$. No per-objective structure survives past the reward.

\paragraph{Advantage-level.}
Per-objective critics $V_\phi^{(k)}(s,\mathbf{w})$ and per-objective GAE are retained, giving $K{+}1$ advantages $A_t^{(0)}, \ldots, A_t^{(K)}$. These are combined \emph{before} clipping:
\[
A_t^{\mathbf{w}} = \sum_{k=0}^{K} c_k(\mathbf{w})\, A_t^{(k)},
\]
and a single clipped surrogate $L^{\mathrm{clip}}(A_t^{\mathbf{w}})$ is applied to the combined advantage.

\paragraph{Loss-level.}
Per-objective critics $V_\phi^{(k)}(s,\mathbf{w})$ and advantages are retained as above, but each objective is clipped \emph{separately} before combination:
\[
L^{\mathrm{loss}} = \sum_{k=0}^{K} c_k(\mathbf{w})\, L^{\mathrm{clip}}\big(A_t^{(k)}\big).
\]
This is the latest point at which the $K{+}1$ objectives are still combined by a simple weighted sum. When these gradients conflict, we additionally consider replacing this weighted sum with gradient-level conflict resolution via PCGrad~\citep{yu2020gradient}, detailed in Appendix~\ref{app:pcgrad}.

\subsection{Steerability Evaluation Suite}
\label{sec:method_steerability}

Aggregate return and Pareto-front coverage alone cannot distinguish a policy that responds correctly to a preference from one that responds diffusely or not at all~\citep{molins2026controllability}. We therefore introduce a complementary suite that isolates, per preference dimension, how strongly the intended objective responds, how much unintended objectives are disturbed, and whether that response is directionally reliable. This adapts rank-based controllability measures previously used on abstract, low-dimensional MORL benchmarks~\citep{molins2026controllability} to embodied, physically simulated continuous control.

\paragraph{Axis sweep.}
For each style dimension $k \in \{1,\ldots,K\}$, we hold a reference preference $\mathbf{w}^0$ fixed in every coordinate but $k$ and sweep $\alpha \in [\alpha_{\min}, \alpha_{\max}]$ across $M$ evenly spaced values:
\[
\mathbf{w}_k^{(i)} = \alpha_i\, \mathbf{e}_k + (1-\alpha_i)\, \frac{\mathbf{w}^0_{-k}}{\sum \mathbf{w}^0_{-k}}, \qquad i = 1,\ldots,M,
\]
where $\mathbf{e}_k$ is the $k$-th one-hot vector and $\mathbf{w}^0_{-k}$ is $\mathbf{w}^0$ with its $k$-th entry zeroed. Each policy is evaluated at every $\mathbf{w}_k^{(i)}$, yielding an achieved objective vector $J(\mathbf{w}_k^{(i)})$ (defined precisely in Appendix~\ref{app:metrics}).

\paragraph{Self-PRR and Cross-PRR.}
The Preference-Reward Responsiveness (PRR) matrix is the finite-difference slope of objective $j$ along the sweep of dimension $k$:
\[
\mathrm{PRR}[j][k] = \frac{1}{M-1}\sum_{i=1}^{M-1} \frac{J_j(\mathbf{w}_k^{(i+1)}) - J_j(\mathbf{w}_k^{(i)})}{\alpha_{i+1} - \alpha_i}.
\]
Self-PRR$_k = \mathrm{PRR}[k][k]$ measures how strongly dimension $k$ responds to its own preference weight. Cross-PRR$_k = \frac{1}{K-1}\sum_{j\neq k} |\mathrm{PRR}[j][k]|$ measures how much that same change disturbs the other $K{-}1$ objectives.

\paragraph{Selectivity.}
Selectivity summarizes the ratio of intended to unintended response across all $K$ dimensions:
\[
\mathrm{Selectivity} = \frac{\overline{\text{Self-PRR}}}{\overline{|\text{Cross-PRR}|}}.
\]
Selectivity $>1$ indicates a preference change moves its own objective more than it disturbs the others on average.

\paragraph{Rank Accuracy.}
Self-PRR and Cross-PRR are sensitive to noise in the finite-difference slope. Rank Accuracy instead asks a coarser, more robust question: does the \emph{ordering} of preference weights along a sweep predict the \emph{ordering} of achieved reward? For each dimension $k$, we compute the fraction of sweep-point pairs $(a,b)$ for which this holds:
\[
\mathrm{RankAcc}_k = \frac{1}{\binom{M}{2}} \sum_{a<b} \mathbbm{1}
\big[(\alpha_b - \alpha_a)\big(J_k(\mathbf{w}_k^{(b)}) - J_k(\mathbf{w}_k^{(a)})\big) > 0\big].
\]

Together, Self-PRR, Cross-PRR, Selectivity, and Rank Accuracy answer the three questions posed in Section~\ref{sec:intro}: whether a preference change produces the intended response, whether it leaks into unrelated behaviors, and whether that response is directionally reliable.

%===============================================================================

\section{Experiments}

\subsection{ProtoMotions: G1 Humanoid Preference-Conditioned Locomotion}
\label{sec:protomotions}

\paragraph{Setup.}
We evaluate four aggregation strategies, reward-level, advantage-level, loss-level, and PCGrad (Section~\ref{sec:method_aggregation}), on a G1 humanoid trained to walk under a $K{=}5$-dimensional style preference (energy, arm swing, foot clearance, step width, smoothness) alongside a fixed-weight heading-tracking task reward. Preferences are sampled from $\mathrm{Dirichlet}(0.5)$ during training. Each method is trained for $1.2\times10^{9}$ environment steps with three independent seeds; full task and reward definitions are given in Appendix~\ref{app:protomotions_setup}.

\paragraph{Evaluation protocol.}
We measure \emph{steering ability} (Section~\ref{sec:method_steerability}) along the five axes from Section~\ref{sec:intro}: responsiveness (Hypervolume, Expected Utility, Self-PRR), monotonicity (Rank Accuracy), selectivity (Selectivity, Cross-PRR), task preservation (mean task reward $J_{\text{task}}$ for tracking the fixed heading and speed target), and failure behavior (fall rate). These are computed from a per-dimension preference axis sweep together with $N{=}50$ preferences sampled from the training-time $\mathrm{Dirichlet}(0.5)$ distribution, under a fixed-horizon convention that zeroes reward after a fall. Full metric definitions and evaluation hyperparameters are given in Appendix~\ref{app:metrics}.

\paragraph{Main results.}
Table~\ref{tab:protomotions_main} reports every metric as mean $\pm$ std over three training seeds. No single method dominates. Reward-level attains roughly $3\times$ the hypervolume, $2\times$ the self-responsiveness, and roughly $2\times$ the $J_{\text{task}}$ of the other three methods, with error bars that do not overlap across seeds.
 It is, however, the weakest of the four on precision, with the lowest selectivity and the highest cross-dimension leakage. Advantage-level shows the opposite profile: the smallest controllable region but the cleanest separation between preference dimensions. Loss-level and PCGrad fall between the two on both axes. We hypothesize this pattern reflects how much per-objective structure survives before aggregation. Reward-level collapses all $K$ objectives into a single scalar reward before the critic or the advantage estimator ever sees them, leaving no mechanism downstream that could prevent one objective's gradient from dominating or leaking into the others. Advantage-level, loss-level, and PCGrad instead keep per-objective critics and advantages further into the update (Section~\ref{sec:method_aggregation}), which appears to trade raw responsiveness for cleaner separation. Aggregating early thus favors a wide controllable range at the cost of precision, while aggregating late favors precise, disentangled control at the cost of range.

\begin{table*}[h]
\centering
\caption{Main comparison on ProtoMotions (G1 humanoid), mean $\pm$ std over 3 training seeds at 1.2B environment steps. Column groups correspond to the five evaluation axes in Section~\ref{sec:intro}: responsiveness, monotonicity, selectivity, task preservation, and failure behavior. Bold marks the best value per column; $\uparrow$/$\downarrow$ indicate the preferred direction.}
\label{tab:protomotions_main}
\scriptsize
\resizebox{\textwidth}{!}{%
\begin{tabular}{l ccc c cc c c}
\toprule
% & \multicolumn{3}{c}{Responsiveness} & & \multicolumn{2}{c}{Selectivity} & & \\
% \cmidrule(lr){2-4} \cmidrule(lr){6-7}
& \multicolumn{3}{c}{Responsiveness} & Monotonicity & \multicolumn{2}{c}{Selectivity} & Task preservation & Failure behavior \\
\cmidrule(lr){2-4} \cmidrule(lr){5-5} \cmidrule(lr){6-7} \cmidrule(lr){8-8} \cmidrule(lr){9-9}
Method & Hypervolume$\uparrow$ & Exp.\ Utility$\uparrow$ & Self-PRR$\uparrow$ & Rank acc.$\uparrow$ & Selectivity$\uparrow$ & Cross-PRR$\downarrow$ & $J_{\mathrm{task}}\uparrow$ & Fall rate$\downarrow$ \\
\midrule
Reward-level     & \textbf{0.316$\pm$.016} & \textbf{0.763$\pm$.007} & \textbf{0.454$\pm$.004} & \textbf{0.847$\pm$.018} & 1.61$\pm$.11 & 0.285$\pm$.019 & \textbf{0.761$\pm$.014} & \textbf{7.4$\pm$.1\%} \\
Advantage-level   & 0.109$\pm$.011 & 0.568$\pm$.014 & 0.232$\pm$.045 & 0.811$\pm$.001 & \textbf{2.16$\pm$.15} & \textbf{0.108$\pm$.013} & 0.409$\pm$.019 & 8.0$\pm$.3\% \\
Loss-level        & 0.103$\pm$.007 & 0.572$\pm$.010 & 0.225$\pm$.004 & 0.792$\pm$.033 & 2.04$\pm$.06  & 0.112$\pm$.006 & 0.414$\pm$.017 & 7.9$\pm$.1\% \\
PCGrad            & 0.104$\pm$.016 & 0.558$\pm$.029 & 0.233$\pm$.004 & 0.800$\pm$.047 & 2.07$\pm$.17 & 0.114$\pm$.008 & 0.393$\pm$.027 & 8.4$\pm$.1\% \\
\bottomrule
\end{tabular}%
}
\end{table*}

\begin{figure}[h]
\centering
\includegraphics[width=\textwidth]{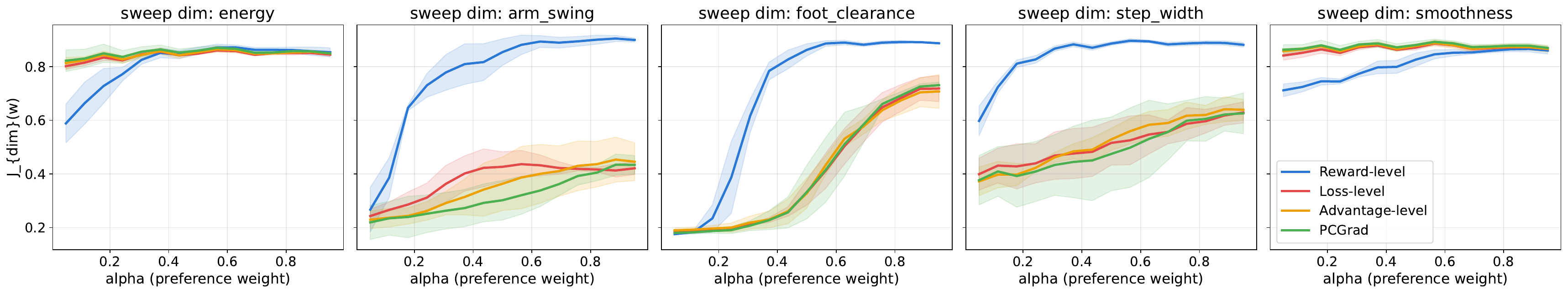}
\caption{Per-dimension preference sweep. Each panel shows the normalized reward achieved for one style dimension as its preference weight increases, mean $\pm$ std shading over 3 training seeds.}
\label{fig:prr_sweep}
\end{figure}

\paragraph{Controllability analysis.}
Figure~\ref{fig:prr_sweep} plots the per-dimension sweep curves underlying Self-PRR: the normalized reward $J_k$ achieved for style dimension $k$ as its preference weight increases from 0.05 to 0.95, averaged over the three seeds. Reward-level's curves sit above and saturate earlier than the other three methods on four of the five style dimensions, consistent with its higher Self-PRR in Table~\ref{tab:protomotions_main}. The gap is not uniform across dimensions: on energy, all four methods converge by $\alpha{\approx}0.4$, suggesting the trade-off we report is driven more strongly by some style dimensions than others.

\subsection{MimicKit: Cross-Framework Validation}
\label{sec:mimickit}

\paragraph{Setup and evaluation.}
To test cross-framework consistency under a fixed movement direction, we repeat reward-, advantage-, and loss-level aggregation in MimicKit using its native optimization and reward processing. The humanoid tracks target speeds sampled from $[0.5,2.0]$ m/s along a fixed $+x$ direction, without a facing reward. The five style objectives and episode-level $\mathrm{Dirichlet}(0.5)$ sampling match the preceding experiment; task and AMP rewards retain fixed coefficients outside the preference vector. Each method is trained for $1.2\times10^{9}$ environment steps with three seeds. We reuse the fixed-horizon protocol with 15-point axis sweeps, $N{=}50$ sampled preferences, 300-step rollouts, and three deterministic evaluation seeds. Table~\ref{tab:mimickit_main} reports mean $\pm$ std across training seeds; Appendix~\ref{app:mimickit} details the setup and native fall criterion.

\paragraph{Results.}
Reward-level achieves the highest mean Self-PRR ($0.510$) and task return, although the Self-PRR differences are modest relative to seed variability. Advantage-level leads in hypervolume and expected utility while minimizing Cross-PRR and falls; loss-level has the highest selectivity, with Rank Accuracy nearly tied between the two later aggregation methods. Thus, later aggregation improves coverage and selectivity here, whereas reward-level has the coverage advantage in ProtoMotions. Reduced cross-objective leakage with later aggregation is shared across both frameworks, but their coverage and utility rankings differ. Reactive-steering results in Appendix~\ref{app:mimickit_steering} show the same precision advantage for later aggregation.

A complementary pattern appears in the MetaDrive ablation: advantage-level aggregation achieves the highest paired expected utility and Self-PRR, whereas loss-level aggregation yields the lowest Cross-PRR (Appendix~\ref{sec:metadrive}, Table~\ref{tab:metadrive_ablation}). This further suggests that later aggregation can improve preference control, although the metric-optimal aggregation level remains environment dependent.

\begin{table*}[h]
\centering
\caption{Fixed-direction preference-conditioned control in MimicKit, mean $\pm$ std over 3 training seeds at 1.2B environment steps. Column groups correspond to the five evaluation axes in Section~\ref{sec:intro}. Bold marks the best displayed mean, including ties; $\uparrow$/$\downarrow$ indicate the preferred direction.}
\label{tab:mimickit_main}
\scriptsize
\resizebox{\textwidth}{!}{%
\begin{tabular}{l ccc c cc c c}
\toprule
& \multicolumn{3}{c}{Responsiveness} & Monotonicity & \multicolumn{2}{c}{Selectivity} & Task preservation & Failure behavior \\
\cmidrule(lr){2-4} \cmidrule(lr){5-5} \cmidrule(lr){6-7} \cmidrule(lr){8-8} \cmidrule(lr){9-9}
Method & Hypervolume$\uparrow$ & Exp.\ Utility$\uparrow$ & Self-PRR$\uparrow$ & Rank acc.$\uparrow$ & Selectivity$\uparrow$ & Cross-PRR$\downarrow$ & $J_{\mathrm{task}}\uparrow$ & Fall rate$\downarrow$ \\
\midrule
Reward-level    & 0.238$\pm$.029 & 0.769$\pm$.009 & \textbf{0.510$\pm$.021} & 0.924$\pm$.003 & 1.272$\pm$.055 & 0.402$\pm$.022 & \textbf{0.870$\pm$.004} & 0.34$\pm$.08\% \\
Advantage-level & \textbf{0.298$\pm$.015} & \textbf{0.776$\pm$.007} & 0.485$\pm$.022 & \textbf{0.937$\pm$.010} & 1.387$\pm$.101 & \textbf{0.350$\pm$.010} & 0.865$\pm$.006 & \textbf{0.25$\pm$.06\%} \\
Loss-level      & 0.290$\pm$.012 & 0.771$\pm$.001 & 0.500$\pm$.030 & \textbf{0.937$\pm$.022} & \textbf{1.422$\pm$.082} & 0.352$\pm$.006 & 0.858$\pm$.005 & 0.32$\pm$.09\% \\
\bottomrule
\end{tabular}%
}
\end{table*}

\section{Limitations and Future Work}
Our study is limited to two simulated humanoid environments with a narrow range of tasks and a small set of hand-designed preference dimensions, and no strategy resolves the responsiveness-precision trade-off we observe. Future work should investigate adaptive objective-combination methods that explicitly balance these properties, other gradient-conflict-resolution methods beyond PCGrad, robustness to out-of-distribution and dynamically changing preferences, and transfer to real robotic systems.

%===============================================================================

\section{Conclusion}
We study when a preference-conditioned policy is truly steerable: whether changing its preference input changes the intended behavior selectively, without compromising the underlying task. Because standard MORL metrics such as hypervolume and expected utility do not capture these properties, we introduce an evaluation suite based on the Preference–Reward Responsiveness matrix, including Self-PRR, Cross-PRR, Selectivity, and Rank Accuracy alongside utility and failure metrics.

Across two humanoid-control frameworks, we show that the level of objective-signal aggregation shapes the trade-off between responsiveness and cross-objective interference. In ProtoMotions, reward-level aggregation achieves the widest controllable range and highest utility but greater leakage, while later aggregation and PCGrad improve selectivity at the cost of range. In MimicKit, all aggregation levels steer reliably, with later aggregation improving coverage or precision without materially reducing task performance. Overall, no fixed strategy dominates, motivating both multidimensional steerability evaluation and adaptive aggregation methods.

\clearpage

\label{sec:conclusion}

\bibliographystyle{plainnat}
\bibliography{reference}  % .bib
\newpage
\appendix
\section{Evaluation Metrics}
\label{app:pareto_metrics}

\begin{definition}[Hypervolume Indicator]
Given a reference point $r \in \mathbb{R}^K$ dominated by every point in a finite set $P = \{p^1,\ldots,p^N\} \subset \mathbb{R}^K$, the hypervolume is
\[
\mathrm{HV}(P; r) = \Lambda\Big(\bigcup_{i=1}^N [r, p^i]\Big),
\]
where $[r,p^i]$ denotes the hyperrectangle spanning from $r$ to $p^i$ and $\Lambda(\cdot)$ is the Lebesgue measure~\citep{hayes2022practical,felten2023toolkit}.
\end{definition}

\begin{definition}[Expected Utility]
Let $\mathcal{W}$ be a distribution over preference vectors and $\pi_w$ the policy conditioned on $w\sim\mathcal{W}$. The expected utility is
\[
\mathrm{EU} = \mathbb{E}_{w\sim\mathcal{W}}\big[w^\top J^{\pi_w}\big].
\]

\end{definition}

In each environment, we instantiate $P$ as the set of $N{=}50$ achieved objective vectors $J(w^{(1)}), \ldots, J(w^{(N)})$ obtained under preferences $w^{(i)}$ sampled from $\mathrm{Dirichlet}(0.5)$, matching the training-time preference distribution. The reference point $r$ is the origin of the normalized $[0,1]^K$ objective space. Hypervolume is estimated by Monte Carlo sampling,
\[
\mathrm{HV} = \Pr_{u \sim \mathcal{U}([0,1]^K)}\Big[\exists\, i : J(w^{(i)}) \succeq u\Big],
\]
using $2\times10^5$ uniform samples $u$. Expected Utility is estimated as the sample mean over the same $N$ points,
\[
\mathrm{EU} = \mathbb{E}_{w \sim \mathrm{Dirichlet}(0.5)}\big[w^\top J(w)\big] \approx \frac{1}{N}\sum_{i=1}^{N} w^{(i)\top} J(w^{(i)}),
\]
following the Expected Utility formulation used in prior preference-conditioned MORL evaluation~\citep{d3po2026,molins2026controllability}.

\section{PCGrad Aggregation}
\label{app:pcgrad}

PCGrad starts from the same $K{+}1$ per-objective clipped losses $\ell_k = c_k(\mathbf{w})\, L^{\mathrm{clip}}(A_t^{(k)})$ as loss-level, but replaces their weighted sum with gradient-level conflict resolution \citep{yu2020gradient}. Each per-objective gradient $g_k = \nabla_\theta \ell_k$ is projected away from any other objective's gradient it conflicts with:
\[
\tilde{g}_i = g_i - \sum_{j:\, g_i \cdot g_j < 0} \frac{g_i \cdot g_j}{\lVert g_j \rVert^2}\, g_j,
\]
using each $g_j$'s original, pre-projection direction. The final actor gradient is $g_{\mathrm{final}} = \sum_{k=0}^{K} \tilde{g}_k$, plus auxiliary gradients from the entropy and bound losses. Non-conflicting objective pairs ($g_i \cdot g_j \geq 0$) are left untouched.

For each outer objective $i \in \{0, \ldots, K\}$, we traverse the other $K$ objectives in an independently sampled random order (\texttt{torch.randperm}), rather than a single shared shuffle order for the entire update step. Each objective's gradient $g_i$ is compared only against the \emph{original}, pre-projection gradients $g_j$ of the objectives it is checked against, so a correction already applied to $g_i$ earlier in its own traversal does not affect how later projections are computed; $i$ is explicitly excluded from its own inner loop to avoid a projected gradient later spuriously conflicting with its own untouched original direction. These choices follow the original PCGrad formulation~\citep{yu2020gradient}.

\section{ProtoMotions Experimental Details}
\label{app:protomotions}

\subsection{Task and Reward Setup}
\label{app:protomotions_setup}

All ProtoMotions experiments train a G1 humanoid to walk using PrefAMP, a preference-conditioned adversarial motion prior. The reward is a weighted sum of one fixed-weight task term and $K{=}5$ preference-weighted style terms:
\[
r = w_{\text{task}}\, r_{\text{heading}} + w_{\text{pref}} \sum_{k=1}^{K} w_k\, r_k, \qquad w_{\text{task}} = 0.3,\ w_{\text{pref}} = 1.0,
\]
where $r_{\text{heading}}$ rewards tracking a target heading velocity and $w = (w_1,\dots,w_K)$, $\sum_k w_k = 1$, is the style preference vector. During training, $w$ is resampled each episode from $\mathrm{Dirichlet}(0.5)$. Table~\ref{tab:app_bounds} lists the five style dimensions and the bounds used to normalize each raw reward to $[0,1]$ before it enters any of the metrics in Appendix~\ref{app:metrics}.

\begin{table}[h]
\centering
\caption{Style reward dimensions and normalization bounds.}
\label{tab:app_bounds}
\begin{tabular}{llc}
\toprule
Dimension & Description & Bounds \\
\midrule
energy          & negative raw power (W)                     & $(-800.0,\ 0.0)$ \\
arm\_swing      & shoulder-pitch angular velocity (rad/s)     & $(0.0,\ 3.0)$ \\
foot\_clearance & swing-phase foot height (m)                 & $(0.0,\ 0.20)$ \\
step\_width     & lateral foot separation (m)                 & $(0.0,\ 0.40)$ \\
smoothness      & negative $\|\Delta\text{action}\|_2$         & $(-2.5,\ 0.0)$ \\
\bottomrule
\end{tabular}
\end{table}

Each episode has a fixed horizon of $H{=}300$ steps. Each of the four aggregation methods (Section~\ref{sec:method_aggregation}) is trained independently for $1.2\times10^{9}$ environment steps ($4096$ parallel environments, $16$-step PPO rollouts) with three random seeds ($42$, $43$, $44$), giving $12$ trained checkpoints in total.

\subsection{Evaluation Metrics and Protocol}
\label{app:metrics}

\paragraph{Fixed-horizon return.}
For a held-fixed preference $w$, we roll out the policy for $H$ steps and compute, for each style dimension $k$,
\[
J_k(w) = \mathbb{E}_{\text{env}}\left[\frac{1}{H}\sum_{t=1}^{H} r_{k,t}^{\text{norm}} \cdot \mathbbm{1}[\text{alive at } t]\right],
\]
where $r_{k,t}^{\text{norm}} \in [0,1]$ is the bounds-normalized reward at step $t$ and the alive indicator zeroes every step after the first fall, so a fall is always penalized regardless of when it occurs. $J_{\text{task}}(w)$ is defined identically using $r_{\text{heading}}$ and its own bounds.

\paragraph{Axis sweep.}
For each dimension $k$ and a reference preference $w^0$ (uniform by default), we construct $M{=}15$ sweep points
\[
w_k^{(i)} = \alpha_i\, e_k + (1-\alpha_i)\, \frac{w^0_{-k}}{\sum w^0_{-k}}, \qquad \alpha_i \in [0.05,\ 0.95] \text{ evenly spaced},
\]
where $e_k$ is the $k$-th one-hot vector and $w^0_{-k}$ is $w^0$ with its $k$-th entry zeroed. This gives $K \times M = 75$ points, evaluated as one batched rollout per swept dimension.

\paragraph{Self-PRR, Cross-PRR, Selectivity.}
The preference-reward responsiveness matrix is the finite-difference slope of $J_j$ along the sweep of dimension $k$:
\[
\mathrm{PRR}[j][k] = \frac{1}{M-1}\sum_{i=1}^{M-1} \frac{J_j(w_k^{(i+1)}) - J_j(w_k^{(i)})}{\alpha_{i+1} - \alpha_i}.
\]
Self-PRR$_k = \mathrm{PRR}[k][k]$ measures how strongly dimension $k$ responds to its own preference weight; Cross-PRR$_k = \frac{1}{K-1}\sum_{j\neq k} |\mathrm{PRR}[j][k]|$ measures how much it leaks into the others. Selectivity is the ratio of their means across all $K$ dimensions, $\overline{\text{Self-PRR}} / \overline{|\text{Cross-PRR}|}$.

\paragraph{Rank accuracy.}
For each dimension $k$, Rank Accuracy is the fraction of sweep-point pairs $(a,b)$ for which the ordering of preference weights predicts the ordering of achieved reward:
\[
\text{RankAcc}_k = \frac{1}{\binom{M}{2}} \sum_{a<b} \mathbbm{1}\big[(\alpha_b - \alpha_a)\big(J_k(w_k^{(b)}) - J_k(w_k^{(a)})\big) > 0\big].
\]

\paragraph{Hypervolume and Expected Utility.}
In addition to the $75$ axis-sweep points, we draw $N{=}50$ preference vectors from $\mathrm{Dirichlet}(0.5)$, matching the training-time distribution, and evaluate the resulting objective vectors $J(w^{(1)}),\dots,J(w^{(N)})$ in one batched rollout. Hypervolume is estimated by Monte Carlo sampling with reference point at the origin of the normalized $[0,1]^K$ space:
\[
\mathrm{HV} = \Pr_{u \sim \mathcal{U}([0,1]^K)}\Big[\exists\, i : J(w^{(i)}) \succeq u\Big],
\]
estimated from $2\times10^5$ uniform samples $u$. Expected Utility is
\[
\mathrm{EU} = \mathbb{E}_{w \sim \mathrm{Dirichlet}(0.5)}\big[w^\top J(w)\big] \approx \frac{1}{N}\sum_{i=1}^{N} w^{(i)\top} J(w^{(i)}),
\]
following the definition used in prior preference-conditioned MORL evaluation~\citep{d3po2026,molins2026controllability}.

\paragraph{Statistical methodology.}
Every checkpoint is evaluated with $3$ repeats using a fixed base evaluation seed of $1000$ (i.e.\ eval seeds $1000$, $1001$, $1002$), shared across all four methods and all three training seeds for consistent evaluation across checkpoints. Table~\ref{tab:app_hparams} summarizes all evaluation hyperparameters. ProtoMotions results are computed in two stages: first the mean over the $3$ eval repeats for a given training-seed checkpoint, then the mean and standard deviation (population, $\text{ddof}{=}0$) of those three per-seed means across the $3$ training seeds. Thus, the reported dispersion is across training-seed means after averaging eval-time rollout noise. MimicKit uses sample standard deviation as specified in Appendix~\ref{app:mimickit}.

\begin{table}[h]
\centering
\caption{Evaluation hyperparameters, shared across all four methods and three training seeds.}
\label{tab:app_hparams}
\begin{tabular}{lc}
\toprule
Horizon $H$ & 300 \\
Axis-sweep points per dimension $M$ & 15 \\
Axis-sweep range $[\alpha_{\min}, \alpha_{\max}]$ & $[0.05, 0.95]$ \\
Full-simplex samples $N$ & 50 \\
Dirichlet concentration $\alpha$ & 0.5 \\
Hypervolume Monte Carlo samples & $2\times10^5$ \\
Eval repeats & 3 (seeds $1000/1001/1002$) \\
\bottomrule
\end{tabular}
\end{table}

\subsection{Per-Dimension Breakdown}
\label{app:per_dim}

Table~\ref{tab:app_per_dim} decomposes Self-PRR, Cross-PRR, and Rank Accuracy by style dimension. The trade-off summarized in Table~\ref{tab:protomotions_main} is not uniform across dimensions: reward-level's advantage in Self-PRR is driven primarily by arm\_swing and foot\_clearance, while on energy it is the weakest of the four methods, consistent with the near-convergence of all four curves on that dimension in Figure~\ref{fig:prr_sweep}.

\begin{table}[h]
\centering
\caption{Per-dimension Self-PRR, Cross-PRR, and Rank Accuracy, mean $\pm$ std over 3 training seeds.}
\label{tab:app_per_dim}
\small
\begin{tabular}{llccc}
\toprule
Method & Dimension & Self-PRR$\uparrow$ & Cross-PRR$\downarrow$ & Rank acc.$\uparrow$ \\
\midrule
\multirow{5}{*}{Reward-level}
& energy          & 0.295$\pm$.051 & 0.441$\pm$.042 & 0.791$\pm$.059 \\
& arm\_swing      & 0.703$\pm$.080 & 0.152$\pm$.044 & 0.895$\pm$.054 \\
& foot\_clearance & 0.790$\pm$.003 & 0.290$\pm$.035 & 0.876$\pm$.026 \\
& step\_width     & 0.315$\pm$.043 & 0.167$\pm$.040 & 0.778$\pm$.003 \\
& smoothness      & 0.165$\pm$.010 & 0.375$\pm$.059 & 0.893$\pm$.028 \\
\midrule
\multirow{5}{*}{Loss-level}
& energy          & 0.048$\pm$.010 & 0.067$\pm$.018 & 0.678$\pm$.035 \\
& arm\_swing      & 0.198$\pm$.002 & 0.040$\pm$.011 & 0.746$\pm$.156 \\
& foot\_clearance & 0.594$\pm$.047 & 0.336$\pm$.017 & 0.984$\pm$.003 \\
& step\_width     & 0.256$\pm$.019 & 0.063$\pm$.024 & 0.939$\pm$.017 \\
& smoothness      & 0.027$\pm$.014 & 0.051$\pm$.013 & 0.614$\pm$.042 \\
\midrule
\multirow{5}{*}{Advantage-level}
& energy          & 0.040$\pm$.013 & 0.060$\pm$.011 & 0.659$\pm$.047 \\
& arm\_swing      & 0.240$\pm$.085 & 0.038$\pm$.011 & 0.944$\pm$.013 \\
& foot\_clearance & 0.575$\pm$.064 & 0.331$\pm$.034 & 0.981$\pm$.003 \\
& step\_width     & 0.297$\pm$.084 & 0.076$\pm$.010 & 0.938$\pm$.013 \\
& smoothness      & 0.009$\pm$.007 & 0.037$\pm$.011 & 0.532$\pm$.034 \\
\midrule
\multirow{5}{*}{PCGrad}
& energy          & 0.030$\pm$.039 & 0.056$\pm$.041 & 0.585$\pm$.141 \\
& arm\_swing      & 0.238$\pm$.043 & 0.041$\pm$.006 & 0.973$\pm$.009 \\
& foot\_clearance & 0.612$\pm$.014 & 0.347$\pm$.022 & 0.979$\pm$.014 \\
& step\_width     & 0.277$\pm$.030 & 0.086$\pm$.019 & 0.923$\pm$.020 \\
& smoothness      & 0.007$\pm$.011 & 0.042$\pm$.024 & 0.540$\pm$.064 \\
\bottomrule
\end{tabular}
\end{table}

\subsection{Measurement Validity: Fixed vs.\ Alive Convention}
\label{app:fixed_vs_alive}

All results in the main text use the fixed-horizon convention above, which divides by the full horizon $H$ and therefore always penalizes a fall. As a robustness check, we also compute a \emph{quality-while-alive} statistic that does not penalize falls in the same way, a pooled ratio of accumulated reward over accumulated alive steps across all environments:
\[
J_k^{\text{alive}}(w) = \frac{\sum_{\text{env}} \sum_{t} r_{k,t}^{\text{norm}} \cdot \mathbb{1}[\text{alive at } t]}{\sum_{\text{env}} (\text{alive steps})}.
\]
This answers a different question, how good the behavior is conditional on survival, rather than whether the preference is practically usable end to end, and the two are not on the same numeric scale. Table~\ref{tab:app_fixed_alive} reports Hypervolume and Expected Utility under both conventions. As expected, Alive values are uniformly higher than Fixed values, since removing the fall penalty can only increase the statistic. More importantly, every best-in-column method in Table~\ref{tab:protomotions_main} remains the best under the Alive convention: reward-level remains highest on Hypervolume, Expected Utility, Self-PRR, and Rank Accuracy, and advantage-level remains highest on Selectivity and lowest on Cross-PRR, under both conventions. The trade-off we report is therefore not an artifact of how failures are scored.

\begin{table}[h]
\centering
\caption{Hypervolume and Expected Utility under the Fixed and Alive conventions, mean $\pm$ std over 3 training seeds.}
\label{tab:app_fixed_alive}
\begin{tabular}{lcccc}
\toprule
& \multicolumn{2}{c}{Hypervolume} & \multicolumn{2}{c}{Expected Utility} \\
\cmidrule(lr){2-3} \cmidrule(lr){4-5}
Method & Fixed & Alive & Fixed & Alive \\
\midrule
Reward-level     & 0.316$\pm$.016 & 0.435$\pm$.021 & 0.763$\pm$.007 & 0.818$\pm$.008 \\
Advantage-level  & 0.109$\pm$.011 & 0.158$\pm$.012 & 0.568$\pm$.014 & 0.610$\pm$.015 \\
PCGrad           & 0.104$\pm$.016 & 0.154$\pm$.024 & 0.558$\pm$.029 & 0.601$\pm$.031 \\
Loss-level       & 0.103$\pm$.007 & 0.149$\pm$.010 & 0.572$\pm$.010 & 0.615$\pm$.011 \\
\bottomrule
\end{tabular}
\end{table}

As a further sanity check, note that a policy for which the preference had no effect at all (i.e.\ every sampled preference produces the same, near-uniform objective vector around $J_k \approx 0.5$) would have a Hypervolume around $0.5^K \approx 0.03$. Every method's Fixed Hypervolume in Table~\ref{tab:app_fixed_alive} exceeds this degenerate baseline by a factor of $3\times$ to $10\times$, indicating that all four methods produce genuinely preference-responsive behavior rather than a collapsed, preference-insensitive policy.

\section{MimicKit Experimental Details}
\label{app:mimickit}

\subsection{Task and Reward Setup}
\label{app:mimickit_setup}

All MimicKit experiments train the framework's native 28-actuated-DoF humanoid (\texttt{data/assets/humanoid/humanoid.xml}) using PrefAMP in Isaac Lab. The reward is a weighted sum of fixed task and AMP terms and $K{=}5$ preference-weighted style terms:
\[
r = w_{\mathrm{task}}\,r_{\mathrm{task}} + w_{\mathrm{AMP}}\,r_{\mathrm{AMP}}
+ w_{\mathrm{pref}}\sum_{k=1}^{K} w_k\,\tilde r_k,
\]
with
\[
w_{\mathrm{task}}=0.1875,\qquad w_{\mathrm{AMP}}=0.1875,\qquad
w_{\mathrm{pref}}=0.625.
\]
Here, $\tilde r_k\in[0,1]$ is normalized style reward $k$ and $w=(w_1,\dots,w_K)$, $\sum_k w_k=1$, is the style preference vector. These coefficients are the sum-to-one normalization of the intended task:AMP:style ratio $0.3{:}0.3{:}1.0$. During training, $w$ is resampled each episode from $\mathrm{Dirichlet}(0.5)$ and supplied to both actor and critic. The five style dimensions and normalization bounds match Table~\ref{tab:app_bounds}.

For the fixed-direction experiment in Table~\ref{tab:mimickit_main}, the task reward is
\[
r_{\mathrm{task}} = r_{\mathrm{velocity}}.
\]
The velocity term exponentially penalizes target-velocity error with scale $0.5$ and is zero for motion opposite the target direction. The target direction remains fixed along the world $+x$ axis, while target speed is sampled from $[0.5,2.0]$ m/s and resampled every 4--7 seconds. The facing reward and facing-direction observation are disabled. This matches ProtoMotions in keeping the movement direction fixed; the speed-command schedule, robot, motion data, and native optimization remain specific to MimicKit.

Position targets are applied at 30 Hz with simulation at 120 Hz. Each episode has a fixed horizon of $H{=}300$ control steps (10 seconds). The AMP discriminator uses \texttt{dataset\_humanoid\_locomotion.yaml}, which contains 56 forward, left, and right running and walking clips. The robot and motion data therefore remain framework-specific, whereas the style dimensions and evaluation protocol match ProtoMotions.

Each of the three aggregation methods (Section~\ref{sec:method_aggregation}) is trained independently for $1.2\times10^{9}$ environment steps ($4096$ parallel environments, $32$-step PPO rollouts) with three random seeds ($20260827$, $20260828$, $20260829$), giving nine fixed-direction checkpoints. The additional reactive-steering experiments are described in Appendix~\ref{app:mimickit_steering}.

\subsection{Optimization and Evaluation}

Table~\ref{tab:mimickit_hparams} summarizes the optimization hyperparameters shared across all methods and seeds.

\begin{table}[h]
\centering
\caption{MimicKit optimization hyperparameters, shared across all methods and seeds.}
\label{tab:mimickit_hparams}
\small
\begin{tabular}{lc}
\toprule
Setting & Value \\
\midrule
Parallel training environments & 4096 \\
PPO rollout length & 32 steps \\
Actor / critic / discriminator hidden widths & 1024 / 1024 / 1024\\
Gaussian action standard deviation & 0.05 \\
Discount $\gamma$ / GAE $\lambda$ & $0.99$ / $0.95$ \\
PPO clipping coefficient & $0.2$ \\
Actor / critic / discriminator learning rate & $5\!\times\!10^{-5}$ / $10^{-4}$ / $2.5\!\times\!10^{-4}$ \\
Actor / critic / discriminator update epochs & 5 / 2 / 2 \\
\bottomrule
\end{tabular}
\end{table}

The normalized returns and PRR-derived metrics follow Appendix~\ref{app:metrics}. Each checkpoint is evaluated in 256 parallel environments with $H{=}300$, $M{=}15$, $N{=}50$, and deterministic seeds $1000$, $1001$, and $1002$. We first average the three evaluation repeats for each checkpoint, then report the mean and sample standard deviation ($\text{ddof}{=}1$) across the three training seeds.

For both MimicKit task settings, the reported fall rate uses the native environment failure flag, triggered by ground contact from a body other than the feet, with pose-distance termination disabled. Returns are accumulated through the first failure step and zeroed thereafter through the 300-step horizon, without resetting failed environments. No root-height threshold is used for these reported results. ProtoMotions uses a different environment-specific termination rule, so absolute fall rates should not be compared directly across frameworks.

\subsection{Reactive-Steering Task Variant}
\label{app:mimickit_steering}

\paragraph{Setup.}
To assess whether the aggregation comparison extends to changing movement directions, we also evaluate a reactive-steering task. Target directions are sampled uniformly and target speeds from $[0.5,2.0]$ m/s; both are resampled every 4--7 seconds. The task reward is
\[
r_{\mathrm{task}} = 0.7\,r_{\mathrm{velocity}} + 0.3\,r_{\mathrm{facing}},
\]
where $r_{\mathrm{facing}}$ is the nonnegative alignment between the humanoid's facing direction and the target direction. The facing-direction observation is enabled. The robot, motion data, style objectives, outer task/AMP/style coefficients, optimization settings, and evaluation protocol are the same as in the fixed-direction experiment. Each method is trained for $1.2\times10^{9}$ environment steps with seeds $20260817$, $20260818$, and $20260819$, giving nine additional checkpoints. Because the task reward and facing observation also change, this comparison does not isolate the effect of direction changes alone, and $J_{\mathrm{task}}$ should be compared across methods within each task setting.

\paragraph{Results.}
Table~\ref{tab:mimickit_steering} reports the reactive-steering results. The methods have similar mean Self-PRR ($0.471$--$0.480$), with Rank Accuracy of at least $0.966$. Advantage-level has the highest mean expected utility, Self-PRR, and selectivity; loss-level has the highest hypervolume and Rank Accuracy and the lowest Cross-PRR and fall rate. Reward-level retains the highest task return. As in the fixed-direction experiment, both later aggregation methods achieve higher mean hypervolume and selectivity and lower Cross-PRR than reward-level. Their precision advantage therefore persists across the two MimicKit task settings, while the highest mean Self-PRR shifts from reward-level under fixed direction to advantage-level under reactive steering.

\begin{table*}[htbp]
\centering
\caption{Reactive-steering preference-conditioned control in MimicKit, mean $\pm$ std over 3 training seeds at 1.2B environment steps. Bold marks the best mean; $\uparrow$/$\downarrow$ indicate the preferred direction.}
\label{tab:mimickit_steering}
\scriptsize
\resizebox{\textwidth}{!}{%
\begin{tabular}{l ccc c cc c c}
\toprule
& \multicolumn{3}{c}{Responsiveness} & Monotonicity & \multicolumn{2}{c}{Selectivity} & Task preservation & Failure behavior \\
\cmidrule(lr){2-4} \cmidrule(lr){5-5} \cmidrule(lr){6-7} \cmidrule(lr){8-8} \cmidrule(lr){9-9}
Method & Hypervolume$\uparrow$ & Exp.\ Utility$\uparrow$ & Self-PRR$\uparrow$ & Rank acc.$\uparrow$ & Selectivity$\uparrow$ & Cross-PRR$\downarrow$ & $J_{\mathrm{task}}\uparrow$ & Fall rate$\downarrow$ \\
\midrule
Reward-level    & 0.178$\pm$.005 & 0.728$\pm$.006 & 0.479$\pm$.020 & 0.966$\pm$.007 & 1.315$\pm$.034 & 0.365$\pm$.023 & \textbf{0.813$\pm$.010} & 0.26$\pm$.04\% \\
Advantage-level & 0.199$\pm$.004 & \textbf{0.732$\pm$.003} & \textbf{0.480$\pm$.004} & 0.971$\pm$.005 & \textbf{1.503$\pm$.015} & 0.319$\pm$.002 & 0.809$\pm$.003 & 0.22$\pm$.03\% \\
Loss-level      & \textbf{0.201$\pm$.005} & 0.731$\pm$.003 & 0.471$\pm$.013 & \textbf{0.975$\pm$.008} & 1.498$\pm$.067 & \textbf{0.315$\pm$.007} & 0.807$\pm$.010 & \textbf{0.17$\pm$.03\%} \\
\bottomrule
\end{tabular}%
}
\end{table*}

\newpage
\section{MetaDrive}
\label{sec:metadrive}

\paragraph{Setup.}
As an environment ablation, we repeat the aggregation comparison in MetaDrive, a procedurally generated autonomous-driving environment. We compare Reward-level, loss-level and advantage-level. Each policy is conditioned on a $K{=}9$ preference vector over driving objectives including progress, speed, lane keeping, road clearance, heading alignment, and route completion, while MetaDrive's native task reward retains a fixed unit weight. All variants use the same architecture.

\begin{table}[h]
\centering
\caption{MetaDrive environment ablation, mean $\pm$ std over three evaluation seeds. Bold marks the best value; $\uparrow$/$\downarrow$ indicate the preferred direction.}
\label{tab:metadrive_ablation}
\small
% \resizebox{\columnwidth}{!}{%
\begin{tabular}{lccc}
\toprule
Method & Paired EU$\uparrow$ & Mean Self-PRR$\uparrow$ & Mean Cross-PRR$\downarrow$ \\
\midrule
Reward-level & 0.608$\pm$.014 & 0.0028$\pm$.0027 & 0.0216$\pm$.0012 \\
Advantage-level & \textbf{0.634$\pm$.015} & \textbf{0.0263$\pm$.0017} & 0.0246$\pm$.0015 \\
Loss-level       & 0.586$\pm$.020 & 0.0015$\pm$.0003 & \textbf{0.0112$\pm$.0005} \\
% ICA       & 0.471$\pm$.021 & $-$0.0045$\pm$.0046 & 0.0261$\pm$.0014 \\
\bottomrule
\end{tabular}
% }
\end{table}

\end{document}